%% file: main.tex
\pdfoutput=1
\documentclass[11pt]{article}

\usepackage[
  twocolumnmode,
  shownumpages,
  bgcolor={250,248,241},
  braincolor={157,225,252},
  linkcolor={133,76,101},
  citecolor={133,76,101},
  urlcolor={133,76,101},
  citingstyle=authoryear,
  bibliostyle=plainnat,
  bibfile=custom
]{Styles/brainlab}
\usepackage{microtype}
\usepackage{inconsolata}
\usepackage{multirow}
\usepackage{wrapfig}
\input{math_commands.tex}

\allowdisplaybreaks
\usepackage{placeins}
\usepackage{needspace}
\setbrainmeta{
  title={SimplexLoRA: Dynamic Rank Updating via Alternating Minimization with Simplex Projection},
  venue={EMNLP 2026},
  authors={
  \textbf{Igor Shalygin}\textsuperscript{1,2,*},
  \textbf{Grigorii Davydenko}\textsuperscript{1,2,3,*},
  \textbf{Andrey Veprikov}\textsuperscript{1,2,3},
  \textbf{Aleksey Antonov}\textsuperscript{1,2,4}
  \\
  \textbf{Nikolay Bushkov}\textsuperscript{4},
  \textbf{Stanislav Moiseev}\textsuperscript{4},
  \textbf{Aleksandr Beznosikov}\textsuperscript{1,2,5}
  },
  affiliations={
\textsuperscript{1}Moscow Independent Research Institute of Artificial Intelligence (MIRAI)
  \\
  \textsuperscript{2}Basic Research of Artificial Intelligence Laboratory (BRAIn Lab)
  \\
  \textsuperscript{3}SB AI Lab
  \qquad
  \textsuperscript{4}T-Technologies
  \qquad
  \textsuperscript{5}Innopolis University
  \\
  \textsuperscript{*}Equal contribution.
  },
  abstract={
The success of large language models in real-world applications relies on parameter-efficient fine-tuning methods such as Low-Rank Adaptation (\texttt{LoRA}), which inserts compact, trainable adapters into selected layers. In this paper, we propose \texttt{SimplexLoRA}, a \texttt{LoRA} variant that adaptively adjusts adapter ranks during fine-tuning based on their learned importance. Unlike standard \texttt{LoRA}, which employs a fixed rank across all layers, \texttt{SimplexLoRA} optimizes per-adapter importance weights on the $n$-scaled simplex during a short warm-up and then freezes them, dynamically allocating capacity where it is most beneficial without increasing the overall parameter budget or inference cost. We evaluate \texttt{SimplexLoRA} across 12 language understanding, summarization, and reasoning tasks using DeBERTaV3, BART-large, Llama~3.1~8B, and Qwen3-8B. Our results show that \texttt{SimplexLoRA} consistently matches or outperforms standard \texttt{LoRA} and recent adaptive baselines, with the largest gains at low ranks. Our code is available at \url{https://github.com/brain-lab-research/SimplexLoRA}.
  }
}

\begin{document}
\begin{mainpart}

\section{Introduction}\label{introduction}
    \texttt{LoRA} \citep{hu2021lora} is a widely used parameter-efficient fine-tuning method \citep{han2024parameter} that parameterizes weight updates as $\mW' = \mW^0 + \mA\mB$, with $\mW^0 \in \R^{m\times d}$ frozen and trainable $\mA \in \R^{m\times r}, \mB \in \R^{r\times d}$, $r \ll m, d$. This approach achieves low inference overhead and substantially reduces memory during fine-tuning.

    \begin{figure}[t]
        \centering
        \includegraphics[width=0.93\linewidth]{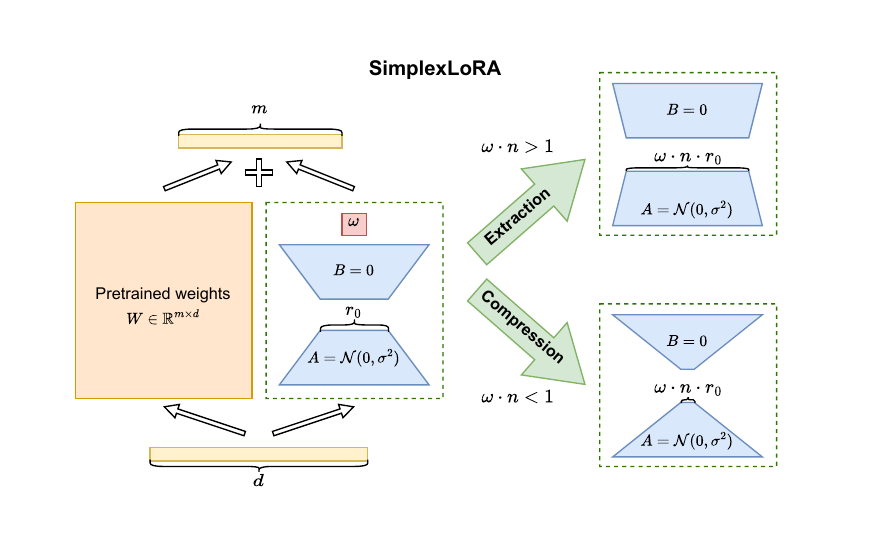}
        \caption{\texttt{SimplexLoRA} framework. Per-adapter importance weights are constrained to the $n$-scaled simplex, enabling dynamic allocation of a fixed rank budget across layers.}
        \label{fig:simplex_lora}
    \end{figure}
    Despite its efficiency, standard \texttt{LoRA} uses a fixed rank across all layers, ignoring layer-wise differences in convergence and representation learning. Existing adaptive rank allocation methods attempt to address this issue but suffer from the critical drawbacks: (i) substantially more trainable parameters \citep{aroma}, (ii) non-negligible inference-time overhead caused by dynamic pruning or gating mechanisms \citep{zhang2025more}, and (iii) deviations from the standard \texttt{LoRA} pipeline that hinder compatibility \citep{zhang2023adalora}. See Section~\ref{sec:related_work} for further discussion.
    
    To address this limitation, we propose \texttt{SimplexLoRA}, which assigns an importance weight to each of the $n$ \texttt{LoRA} adapters, constrained to the $n$-scaled simplex to enforce a fixed parameter budget. Key properties: (i) ranks learned automatically during fine-tuning; (ii) simplex constraint preserves the \texttt{LoRA} parameter budget; (iii) full \texttt{LoRA} compatibility; (iv) negligible training and inference overhead.

\section{Related Work}\label{sec:related_work}

\subsection{Classical low-rank approaches}

While \texttt{LoRA} has popularized parameter-efficient fine-tuning, it is not the only method to explore low-rank or structured adaptations. Earlier approaches include \texttt{LoHA} \citep{hyeon2021fedpara}, which utilizes element-wise (Hadamard) interactions between low-rank matrices to reduce the parameter count without sacrificing expressivity, and \texttt{Krona} \citep{edalati2022krona}, which employs the Kronecker product in place of the conventional low-rank matrix product to improve parameter efficiency while maintaining model capacity.

Subsequent methods extended or refined the \texttt{LoRA} framework. \texttt{VERA} \citep{kopiczko2023vera} reduces adapter storage by sharing frozen random projection matrices across layers and learning small scaling vectors. \texttt{rsLoRA} \citep{kalajdzievski2023rank} replaces the original \texttt{LoRA} scaling factor $\alpha/r$ \citep{hu2021lora} with the rank-stabilized factor $\alpha/\sqrt{r}$. A complementary direction is taken by \texttt{DoRA} \citep{liu2024dora}, which decomposes the adapted weight into magnitude and direction components and learns the magnitude together with a low-rank correction of the direction.

Another line of work improves \texttt{LoRA} optimization and numerical efficiency. \texttt{LoRA+} \citep{hayou2024loraplus} uses different learning rates for the two adapter matrices, while \texttt{QLoRA} \citep{dettmers2023qlora} enables memory-efficient fine-tuning through 4-bit quantization of frozen base-model weights. Both are orthogonal to our adaptive rank allocation method and can be combined with it.

\subsection{Dynamic \texttt{LoRA} rank allocation}

This topic is of particular relevance in the context of our work, as we focus on adaptive rank learning, enabling the model to automatically adjust the capacity of each adapter during fine-tuning. It has attracted significant attention in recent research \citep{increlora,zhang2023adalora,aroma}, as exploiting dynamically changing ranks allows models to utilize newly allocated parameters more efficiently.

The SVD-like representation of \texttt{LoRA} adapters proposed in \texttt{AdaLoRA} \citep{zhang2023adalora} replaces the standard two-matrix adapter with the parameterization $\mP\Lambda\mQ$ and enables the construction of importance scores for individual singular components. Components outside the global importance budget are masked by setting the corresponding entries of $\Lambda$ to zero. To encourage the SVD-like structure, the method applies the orthogonality regularizer
\[
R(\mP,\mQ)
=
\|\mP^\top\mP-\mI\|_F^2
+
\|\mQ^\top\mQ-\mI\|_F^2.
\]
During training, all three matrices $\mP$, $\Lambda$, and $\mQ$ are updated simultaneously. Although this parameterization avoids explicitly computing an SVD at every fine-tuning step, maintaining an initial rank above the target budget, regularly evaluating importance scores, and applying orthogonality regularization increase the training complexity compared with standard \texttt{LoRA}.

Several methods manage the rank budget more flexibly. \texttt{ElaLoRA} \citep{chang2025elalora} uses SVD-based adaptation and gradient-derived importance scores to prune and expand ranks at scheduled intervals. \texttt{FlexLoRA} \citep{liu2026flexlora} uses normalized spectral entropy to guide rank pruning and expansion under a global budget. The dynamic-rank \texttt{DoRA} of \citet{mao2024dora} decomposes adapters into structured rank-one components and prunes them based on per-component importance. \texttt{NormAL LoRA} \citep{aastik2025normallora} uses rank-norm regularization to determine an appropriate rank for each module.

In a similar way, \citet{aroma} proposes a rank-one \texttt{LoRA} series representation of the increment matrix:
\[
\Delta\mW
= \vb_1\va_1+\ldots+\vb_p\va_p
= \begin{bmatrix}\mB_{p-1}&\vb_p\end{bmatrix}
\begin{bmatrix}\mA_{p-1}\\\va_p\end{bmatrix},
\]
where $p$ denotes the number of outer-loop iterations. This representation is applied to several layers, and after each outer-loop step a convergence criterion is evaluated: if the relative change between the $(p-1)$-th and $p$-th solutions is sufficiently small, the training of $\Delta\mW$ is terminated. Unlike \texttt{IncreLoRA}, which expands modules according to gradient-based importance scores, \texttt{AROMA} uses the relative Frobenius-norm change between consecutive solutions as a stopping criterion.

Compared with these works, \texttt{SimplexLoRA} does not force adapter ranks to strictly increase or rely on expensive SVD computations. Instead, we allow the ranks to be learned as continuous parameters, enabling the model to respond dynamically to the task's complexity. By leveraging simplex projection and weight freezing, we obtain a direct form of rank adaptation that avoids explicit importance scoring and iterative pruning.

\section{\texttt{SimplexLoRA} Framework}\label{sec:simplex_lora}

    For the \texttt{LoRA} method, we introduce trainable layer-wise scaling factors:
    $
        \displaystyle \mW_i^0 + \displaystyle \mA_i \displaystyle \mB_i \rightarrow \displaystyle \mW_i^0 + \omega_i \displaystyle \mA_i \displaystyle \mB_i,
    $
    where $\omega_i \geq 0$ represents the importance coefficient assigned to the $i$-th adapter. The weights $\omega = (\omega_1,\dots,\omega_n)$ are constrained to the $n$-scaled simplex, where $n$ is the number of \texttt{LoRA} adapters. This preserves \texttt{LoRA} magnitude, with $\omega = (1,\dots,1)=\mathbf{1}$ recovering standard \texttt{LoRA} initialization~\citep{hu2021lora}.

    Motivated by the theoretical analysis in Appendix~\ref{appendix:theory}, \texttt{SimplexLoRA} formulates the following alternating optimization problem:
    \begin{align}
    \label{eq:simplex_lora}
   \min_{\omega \in n \cdot \Delta_{n-1}} \min_{\mathcal{A}, \mathcal{B}}
    \mathcal{L}(\omega,\mathcal{A},\mathcal{B})
    + (\lambda / 2)\|\omega-\mathbf{1}\|_2^2,
    \end{align}
    where $\mathcal{A}, \mathcal{B} = \{\displaystyle \mA_i \}_{i=1}^n, \{\displaystyle \mB_i \}_{i=1}^n$ denote the sets of all LoRA adapters, $\mathcal{L}$ is the loss, and the term $(\lambda / 2)\|\omega-\mathbf{1}\|_2^2$ prevents overfitting of the importance weights $\omega$.


\subsection{\texttt{SimplexLoRA} Training Process}
    The \texttt{SimplexLoRA} algorithm employs an alternating optimization scheme with three phases, repeated for \(K\) iterations.

\paragraph{Phase 1. Warm-up.}
    For $T$ steps, we perform standard \texttt{LoRA} fine-tuning by jointly optimizing the adapter parameters $\mathcal{A}, \mathcal{B}$ and the importance weights $\omega$ using gradient-based methods.

\paragraph{Phase 2. Simplex projection.}
    The weight vector $\omega$ is projected onto the scaled simplex $n \cdot \Delta_{n-1}$ via Euclidean projection \citep{6976941}:
    $$
        \min_{\hat{\omega} \in n \cdot \Delta_{n-1}} \|\omega - \hat{\omega}\|_2^2.
    $$

    \paragraph{Phase 3. Rank update.}
    Adapter ranks are dynamically resized while preserving the total parameter budget: $r_i^{\text{new}} = \lfloor r_0 \cdot \omega_i \rfloor$, where $r_0$ is the initial (uniform) \texttt{LoRA} rank. When ranks change, the corresponding adapter matrices are resized and reinitialized following the standard \texttt{LoRA} scheme (see Appendix~\ref{appendix:reinit}).



\paragraph{Repeat Phases 1--3 for $\mathbf{K}$ iterations.}

\paragraph{Phase 4. Standard \texttt{LoRA} training with new ranks.}
    After \(K\) repetitions, we fix \(\omega\) to \(\mathbf{1}\) and reinitialize the adapter matrices $\mathcal{A}$ and $\mathcal{B}$ (Appendix~\ref{appendix:reinit}). In this sense Phases~1--3 act as a short rank-search procedure: their role is to determine the per-layer rank allocation, while the warm-up adapter weights are intentionally discarded so that the final stage is functionally identical to standard \texttt{LoRA} with a non-uniform rank profile. Standard training then resumes with these fixed ranks, and any \texttt{LoRA}-based optimization algorithm can be applied.

    \begin{figure}[!htb]
        \centering
        \includegraphics[width=\linewidth]{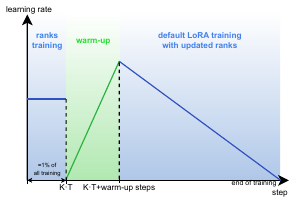}
        \caption{Scheduler for \texttt{SimplexLoRA} parameters.}
        \label{fig:scheduler}
    \end{figure}

\paragraph{Scheduler.}
    To ensure that the model resumes training after freezing the weights, we employ a linear scheduler with two training phases (see Figure \ref{fig:scheduler}). The first phase includes rank training with reduced learning rate. Once rank training is complete, the learning rate is reset to zero, and the warm-up begins to train the resulting \texttt{LoRA} layers. Note, however, that the learning rate for $\omega$ is fixed during training, the scheduler exclusively adjusts the \texttt{LoRA} weights. As a result, the subsequent fine-tuning phase is functionally identical to the standard \texttt{LoRA} training process.

\paragraph{Discussion.}
    The total number of rank updates, \(K \cdot T\), is kept small to avoid rank collapse and to ensure that the majority of training focuses on optimizing the \texttt{LoRA} adapters. In practice, \(T\) is negligible compared to the overall training duration. In the ablation study (Section~\ref{sec:ablation_study}), we analyze the sensitivity of \texttt{SimplexLoRA} to both hyperparameters \(K\) and \(T\), evaluating their impact on performance.

\begin{table*}[t]
    \caption{Comparative performance of "out-of-the-box" algorithms on DeBERTaV3\textsubscript{base}. The best results are shown in \textbf{bold}, and the second-best are \underline{underlined}. All experiment details are provided in Appendix \ref{appendix:exp_comp}.}
    \label{tab:glue_all_adapters}
    \centering
    \resizebox{\linewidth}{!}{
    \begin{tabular}{l|c|ccccccc|c}
    \multirow{2}{*}{\bf Method} & \multirow{2}{*}{\bf \small \# Params} & {\bf MNLI} & {\bf SST-2} & {\bf CoLA} & {\bf QQP } & {\bf QNLI} & {\bf RTE}  & {\bf MRPC}  & {\bf ALL} \\
    ~ & ~ & {Acc} & {Acc} & {Mcc} & {Acc/F1} & {Acc} & {Acc} & {Acc/F1} & {Avg} \\
    \midrule 
    {\texttt{Full Fine-Tuning}} & {184M (100\%)} & {\underline{0.8910}} & {0.9541} & {0.6806} & {0.8962/0.8644} & {\bf 0.9383} & {0.8376} & {0.8848/0.9165} & {0.8689} \\
    \midrule 
    $\texttt{LoRA}_{r=8}$ \citep{hu2021lora} & {442K (0.24\%)} & {0.8797} & {0.9450} & {\bf 0.6913} & {0.8802/0.8437} & {0.9301} & {\bf 0.8448} & {0.8897/\underline{0.9220}} & \underline{0.8696} \\
    {$\texttt{LoHA}_{r = 8}$ \citep{hyeon2021fedpara}} & {884K (0.48\%)} & {0.8560} & {0.9392} & {0.6295} & {0.8674/0.8269} & {0.9085} & {0.8051} & {0.8628/0.9007} & {0.8440} \\
    {$\texttt{Krona}_{r = 32}$ \citep{edalati2022krona}} & {57.6K (0.03\%)} & {0.7587} & {0.9289} & {0.5907} & {0.8275/0.7816} & {0.8605} & {0.7509} & {0.7500/0.8416} & {0.7878} \\
    $\texttt{VERA}_{r=1024}$ \citep{kopiczko2023vera} & {1.64M (0.88\%)} & {0.8355} & {0.9404} & {0.6252} & {0.8592/0.8203} & {0.8975} & {0.7581} & {0.8529/0.8962} & {0.8317} \\
    $\texttt{rsLoRA}_{r=8}$ \citep{kalajdzievski2023rank} & {442K (0.24\%)} & {0.8814} & {\underline{0.9564}} & {0.6414} & {0.8853/0.8506} & {0.9325} & {0.8267} & {0.8750/0.9122} & {0.8624} \\
    $\texttt{DoRA}_{r=8}$ \citep{liu2024dora} & {442K (0.24\%)} & {0.8811} & {0.9473} & {0.6748} & {0.8809/0.8451} & {0.9295} & {0.7978} & {0.8799/0.9139} & {0.8611} \\
    \midrule
    {$\texttt{AdaLoRA}_{r = 8}$ \citep{zhang2023adalora}} & {664K (0.36\%)} & {0.8390} & {0.9392} & {0.6222} & {0.8534/0.8083} & {0.8999} & {0.7834} & {0.8859/0.9178} & {0.8388} \\
    $\texttt{NormAL LoRA}_{r=8}$ \citep{aastik2025normallora} & {442K (0.24\%)} & {0.8814} & {0.9328} & {\underline{0.6857}} & {0.8768/0.8403} & {0.9282} & {0.8292} & {0.8870/0.9195} & {0.8645} \\
    $\texttt{ElaLoRA}_{r=8}$ \citep{chang2025elalora} & {442K (0.24\%)} & {0.8853} & {0.9501} & {0.6638} & {\underline{0.8979}/0.8628} & {0.9146} & {0.8031} & {0.8782/0.9113} & {0.8630} \\
    $\texttt{FlexLoRA}_{r=8}$ \citep{liu2026flexlora} & {664K (0.36\%)} & {0.8711} & {0.9557} & {0.6331
    } & {0.8684/\underline{0.8658}} & {0.9343} & {\underline{0.8392}} & {{\bf 0.9014}/0.9217} & {0.8656} \\
    \midrule
    {$\texttt{SimplexLoRA}_{r=8}$} &  {442K (0.24\%)} & {\bf 0.8984} & {\bf 0.9587} & {0.6572} & {{\bf 0.9124}/{\bf 0.8844}} & {\underline{0.9362}} & {0.8339} & {\underline{0.8987}/{\bf 0.9289}} & {\bf 0.8788} \\
    \bottomrule
    \end{tabular}
    }
    
\end{table*}

\subsection{Motivating Example of Rank Scaling}
    \begin{figure}[!htb]
        \centering
        \includegraphics[width=0.92\linewidth]{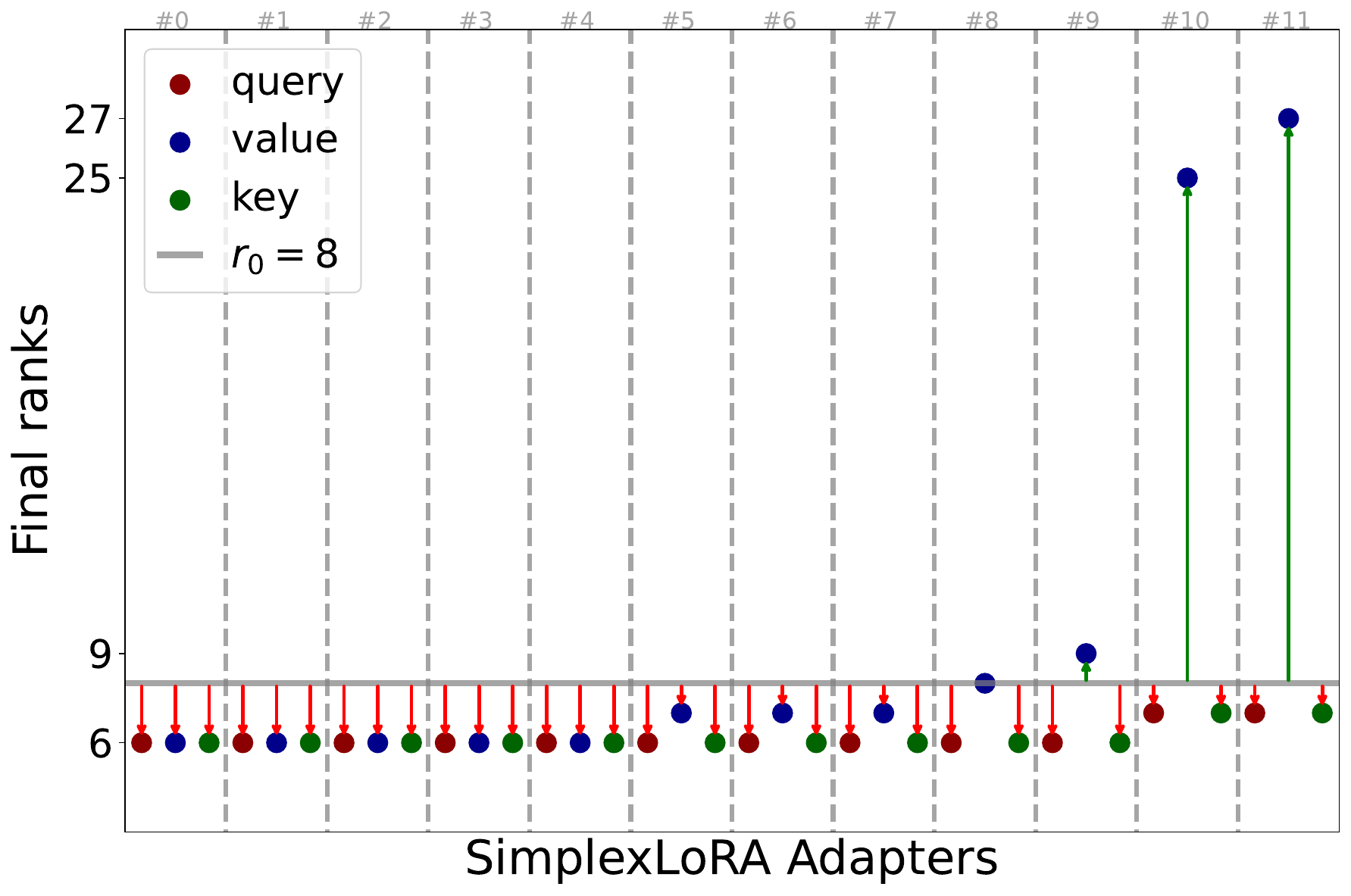}
        \caption{Illustration of the rank distribution of \texttt{SimplexLoRA}.}
        \label{fig:final_ranks}
    \end{figure}

    Figure~\ref{fig:final_ranks} shows the final rank distribution from fine-tuning DeBERTaV3\textsubscript{base} \citep{he2021debertav3} on the MRPC dataset from GLUE \citep{wang2018glue}, tuning only self-attention layers. The learned distribution reveals a clear pattern: value projection matrices in self-attention layers consistently receive the highest ranks, indicating their critical role. 
    Rather than eliminating ranks of adapters with low weights, the method reduces their magnitude, which helps mitigate underfitting.

\section{Experiments}
    We evaluate \texttt{SimplexLoRA} on a diverse set of benchmarks, including GLUE \citep{wang2018glue}, XSum \citep{narayan2018xsum}, CNN/Daily Mail \citep{chen2016cnn}, GSM8K \citep{cobbe2021gsm8k}, BoolQ \citep{clark2019boolq}, and HellaSwag \citep{zellers2019hellaswag}. We exclude GLUE STS-B, whose regression metric requires custom evaluation and which overlaps with the QQP/MRPC similarity tasks. Experiments are conducted using DeBERTaV3\textsubscript{base} \citep{he2021debertav3}, Llama~3.1~8B \citep{grattafiori2024llama3herdmodels}, Qwen3-8B \citep{yang2025qwen3technicalreport}, and BART-large \citep{lewis2019bart} models. Hardware and hyperparameters are provided in Appendix~\ref{appendix:exp_comp}.

\subsection{Out-of-the-box Comparison}
    We benchmark against baselines in default configurations under a number of training iterations. These span foundational PEFT methods (a performance floor) and recent dynamic-rank methods comparable to \texttt{SimplexLoRA}:
    
    \textbf{Standard PEFT methods:} \texttt{LoRA}~\citep{hu2021lora}, \texttt{LoHA}~\citep{hyeon2021fedpara}, \texttt{Krona}~\citep{edalati2022krona}, \texttt{rsLoRA}~\citep{kalajdzievski2023rank}, \texttt{DoRA}~\citep{liu2024dora}, and \texttt{VERA}~\citep{kopiczko2023vera}.
    
    \textbf{Dynamic rank methods:} \texttt{AdaLoRA}~\citep{zhang2023adalora}, \texttt{NormAL LoRA}~\citep{aastik2025normallora}, \texttt{ElaLoRA}~\citep{chang2025elalora}, and \texttt{FlexLoRA}~\citep{liu2026flexlora}.

    In this setting (Table~\ref{tab:glue_all_adapters}), where adapters are applied only to self-attention layers, full fine-tuning, standard \texttt{LoRA}, and \texttt{SimplexLoRA} emerge as the most robust methods, with \texttt{SimplexLoRA} achieving the best average score at the same $0.24\%$ parameter budget as standard \texttt{LoRA}. Other approaches require extensive hyperparameter tuning to achieve competitive performance. Consequently, we select standard \texttt{LoRA} as our primary baseline for further detailed comparison due to its widespread adoption and stability.

\begin{table*}[t]
    \caption{Comparative results of top-performing algorithms: \texttt{LoRA} (L) and \texttt{SimplexLoRA} (S) on DeBERTaV3\textsubscript{base} and Llama 3.1 8B. Results are averaged over 3 runs with different seeds. The best results per rank are shown in \textbf{bold}. All experiment details are provided in Appendix \ref{appendix:exp_comp}.}
    \label{tab:llama_deberta_glue}
    \resizebox{\linewidth}{!}{
    \begin{tabular}{c|c|c|ccccccc|c}    
        \multirow{2}{*}{\bf Model} & \multirow{2}{*}{\bf Rank} & \multirow{2}{*}{\bf Method} & {\bf MNLI} & {\bf SST-2} & {\bf CoLA} & {\bf QQP } & {\bf QNLI} & {\bf RTE}  & {\bf MRPC}  & {\bf ALL} \\
        & & & {Acc} & {Acc} & {Mcc} & {Acc/F1} & {Acc} & {Acc} & {Acc/F1} & {Avg} \\

        \midrule
        
        \multirow{8}*{DeBERTaV3\textsubscript{base}} & \multirow{2}*{$r = 1$} & L & {0.8912} & {0.9488} & {0.6558} & {0.8905/0.8544} & {0.9250} & {0.7943} & {0.8905/0.9213} & {0.8562} \\
        & & S & {\bf 0.8913} & {\bf 0.9576} & {\bf 0.6775} & {{\bf 0.9018}/{\bf 0.8709}} & {\bf 0.9317} & {\bf 0.8159} & {{\bf 0.9093}/{\bf 0.9359}} & {\bf 0.8690} \\
        
        \cmidrule{2-11}
        
        & \multirow{2}*{$r = 2$} & L & {\bf 0.9004} & {0.9413} & {0.6600} & {0.9077/0.8631} & {\bf 0.9369} & {0.8068} & {0.9003/0.9193} & {0.8629} \\
        & & S & {0.8946} & {\bf 0.9608} & {\bf 0.6658} & {{\bf 0.9178}/{\bf 0.8842}} & {0.9338} & {\bf 0.8252} & {{\bf 0.9103}/{\bf 0.9314}} & {\bf 0.8717} \\
        
        \cmidrule{2-11}
        
        & \multirow{2}*{$r = 4$} & L & {0.8780} & {0.9562} & {\bf 0.6809} & {0.9153/0.8866} & {\bf 0.9582} & {0.7816} & {{\bf 0.9207}/{\bf 0.9471}} & {0.8700} \\
        & & S & {\bf 0.9002} & {\bf 0.9610} & {0.6807} & {{\bf 0.9244}/{\bf 0.8903}} & {0.9358} & {\bf 0.8259} & {0.9044/0.9322} & {\bf 0.8756} \\
        
        \cmidrule{2-11}
        
        & \multirow{2}*{$r = 8$} & L & {0.9030} & {\bf 0.9641} & {0.6965} & {0.9092/0.8723} & {\bf 0.9449} & {\bf 0.8537} & {0.9064/0.9127} & {0.8804} \\
        & & S & {\bf 0.9047} & {0.9619} & {\bf 0.6978} & {{\bf 0.9129}/{\bf 0.8901}} & {0.9447} & {0.8305} & {{\bf 0.9267}/{\bf 0.9475}} & {\bf 0.8826} \\
        
        \midrule
    
        \multirow{8}*{Llama 3.1 8B} & \multirow{2}*{$r = 1$} & L & {\bf 0.8961} & {0.9751} & {\bf 0.7015} & {{\bf 0.8891}/0.8370} & {0.9411} & {0.9016} & {0.9154/{\bf 0.9408}} & {0.8886} \\
        & & S & {0.8861} & {\bf 0.9801} & {0.6829} & {{\bf 0.8891}/{\bf 0.8379}} & {\bf 0.9431} & {\bf 0.9204} & {{\bf 0.9278}/0.9404} & {\bf 0.8898} \\
    
        \cmidrule{2-11}
    
        & \multirow{2}*{$r = 2$} & L & {0.8996} & {0.9751} & {0.6967} & {{\bf 0.8851}/0.8311} & {\bf 0.9421} & {0.9154} & {0.9204/0.9441} & {0.8900} \\
        & & S & {\bf 0.8998} & {\bf 0.9801} & {\bf 0.6981} & {{\bf 0.8851}/{\bf 0.8316}} & {0.9381} & {\bf 0.9353} & {{\bf 0.9254}/{\bf 0.9466}} & {\bf 0.8933} \\
        
        \cmidrule{2-11}
    
        & \multirow{2}*{$r = 4$} & L & {0.8993} & {0.9800} & {0.6902} & {0.8851/0.8286} & {0.9391} & {0.9154} & {{\bf 0.9254}/{\bf 0.9462}} & {0.8899} \\
        & & S & {\bf 0.9004} & {\bf 0.9823} & {\bf 0.7069} & {{\bf 0.8891}/{\bf 0.8325}} & {\bf 0.9421} & {\bf 0.9304} & {0.9204/0.9429} & {\bf 0.8941} \\
        
        \cmidrule{2-11}
    
        & \multirow{2}*{$r = 8$} & L &
        {0.8995} & {\bf 0.9851} & {0.6928} & {0.8851/0.8306} & {0.9421} & {0.9104} & {0.9154/0.9419} & {0.8892} \\
        & & S & {\bf 0.9016} & {\bf 0.9851} & {\bf 0.7058} & {{\bf 0.8881}/{\bf 0.8343}} & {\bf 0.9431} & {\bf 0.9353} & {{\bf 0.9204}/{\bf 0.9429}} & {\bf 0.8952} \\
 
        \bottomrule
    \end{tabular}
    }
\end{table*}

\subsection{Comparison with LoRA}
    Table \ref{tab:llama_deberta_glue} compares \texttt{SimplexLoRA} with standard \texttt{LoRA} at multiple ranks on DeBERTaV3\textsubscript{base} and Llama~3.1~8B (adapters in all linear layers, hyperparameters tuned per method). Across most GLUE tasks, \texttt{SimplexLoRA} consistently outperforms \texttt{LoRA} while maintaining the same computational cost. For generality, we also evaluate Qwen3-8B (GSM8K, BoolQ, HellaSwag) and BART-large (XSum, CNN/DailyMail) in Table~\ref{tab:qwen_bart}. As on GLUE, \texttt{SimplexLoRA} matches or outperforms \texttt{LoRA} on most of these tasks, confirming the trend across architectures and task types. The improvement is most pronounced at low ranks ($r \in \{1, 2\}$), where the tight capacity budget of standard \texttt{LoRA} gives adaptive allocation the most room to help.

\begin{table}[!t]
    \caption{Comparative results: \texttt{LoRA} (L) and \texttt{SimplexLoRA} (S) on Qwen3-8B and BART-large. The best results per rank and model are shown in \textbf{bold}. All details are provided in Appendix \ref{appendix:exp_comp}.}
    \label{tab:qwen_bart}
    \centering
    \resizebox{\linewidth}{!}{
    \begin{tabular}{c|c|cccc|cc}
    \multirow{3}{*}{\bf Rank} & \multirow{3}{*}{\bf Method} & \multicolumn{4}{c|}{\bf Qwen3-8B} & \multicolumn{2}{c}{\bf BART-large} \\
    \cmidrule(lr){3-6} \cmidrule(lr){7-8}
    & & {\bf GSM8K} & {\bf BoolQ} & {\bf HellaSwag} & {\bf ALL} & {\bf XSum} & {\bf CNN/DailyMail} \\
    & & {Acc} & {Acc} & {Acc} & {Avg} & {ROUGE-1} & {ROUGE-1} \\
    \midrule

    \multirow{2}*{$r = 1$} & \texttt{L} & {0.8522} & {0.655} & {0.772} & {0.760} & {0.372} & {0.245} \\
    ~ & $\texttt{S}$ & {\bf 0.8666} & {\bf 0.680} & {\bf 0.775} & {\bf 0.774} & {\bf 0.373} & {\bf 0.249} \\
    \midrule

    \multirow{2}*{$r = 2$} & \texttt{L} & {\bf 0.8681} & {0.662} & {0.776} & {0.769} & {0.374} & {0.250} \\
    ~ & $\texttt{S}$ & {0.8673} & {\bf 0.682} & {\bf 0.791} & {\bf 0.780} & {\bf 0.376} & {\bf 0.254} \\
    \midrule

    \multirow{2}*{$r = 4$} & \texttt{L} & {0.8537} & {0.691} & {0.777} & {0.774} & {0.375} & {0.255} \\
    ~ & $\texttt{S}$ & {\bf 0.8719} & {\bf 0.700} & {\bf 0.780} & {\bf 0.784} & {\bf 0.378} & {\bf 0.256} \\
    \midrule

    \multirow{2}*{$r = 8$} & \texttt{L} & {0.8681} & {0.691} & {0.782} & {0.780} & {\bf 0.380} & {0.257} \\
    ~ & $\texttt{S}$ & {\bf 0.8741} & {\bf 0.699} & {\bf 0.790} & {\bf 0.788} & {0.379} & {\bf 0.259} \\
    \bottomrule
    \end{tabular}
    }
    
\end{table}

\Needspace{5\baselineskip}
\subsection{Ablation Study}\label{sec:ablation_study}
    \begin{figure}[!t]
        \centering
        \includegraphics[width=\linewidth]{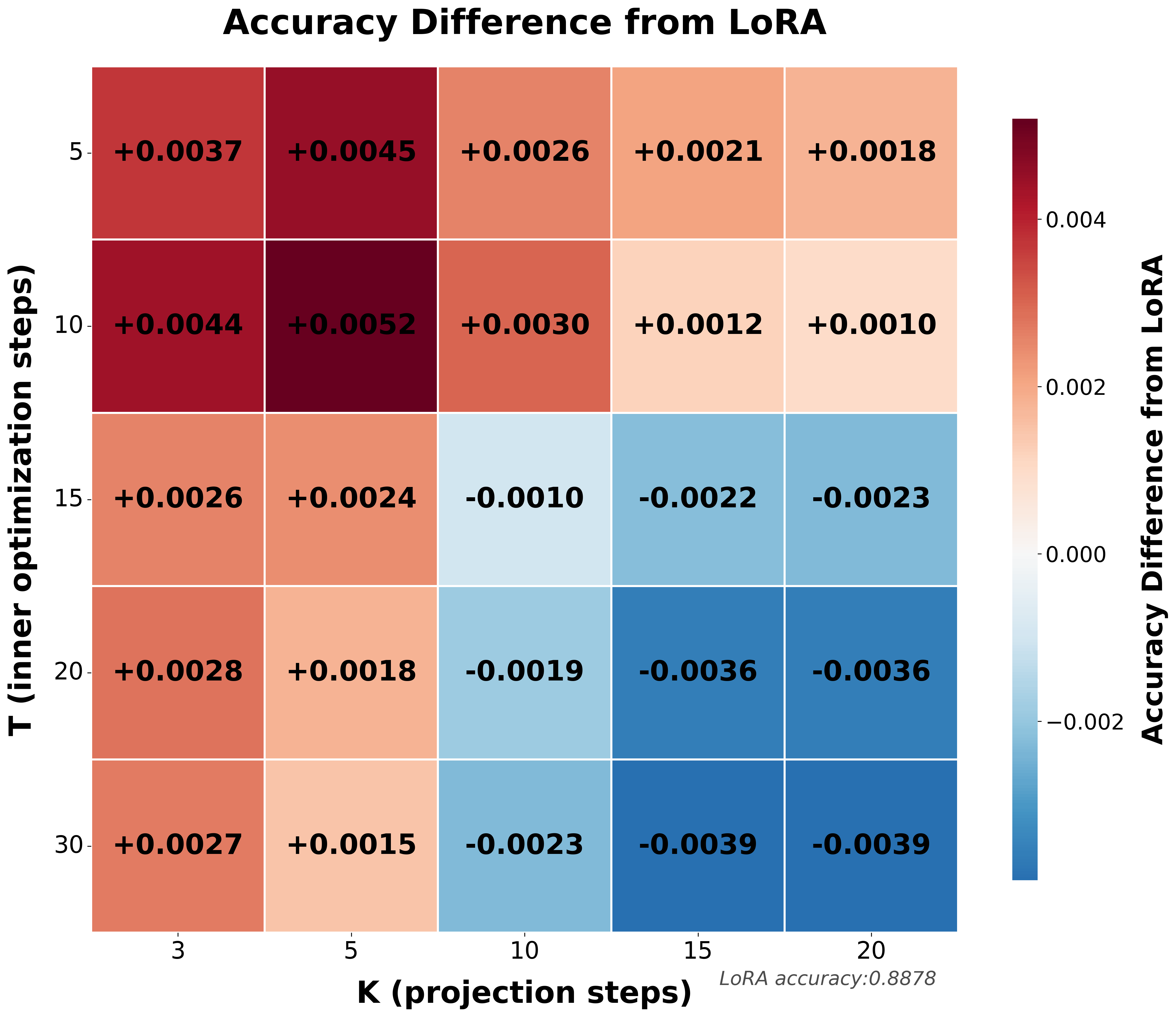}
        \caption{The influence of hyperparameters on \texttt{SimplexLoRA} performance.}
        \label{fig:ablation}
    \end{figure}

    As supplementary analysis, we grid-search hyperparameters on MNLI with DeBERTa to assess the robustness (Figure~\ref{fig:ablation}). Both methods are trained for 2048 steps using the best configuration from the main experiments; each cell reports the validation-accuracy difference.
    
    The results show that \texttt{SimplexLoRA} is stable under a limited optimization budget ($K \cdot T < 100$, i.e., less than 5\% of total training). Very small values (e.g., $K=3$, $T=5$) lead to underfitting due to insufficient rank adaptation, while larger values (e.g., $K\geq10$ or $T\geq15$) tend to cause rank overfitting. Beyond these thresholds, we find that increasing the number of projection steps and inner iterations results in a performance plateau, with values showing minimal deviation from the average. This confirms the importance of keeping $K \cdot T$ low to maintain efficiency without performance loss. In all experiments, we use $K=5$ and $T=10$, which consistently yields strong performance.

\section{Conclusion}
    We presented \texttt{SimplexLoRA}, a \texttt{LoRA} variant that allocates a fixed rank budget non-uniformly across layers via short alternating optimization on the $n$-scaled simplex. Across 12 tasks and four backbones up to 8B parameters, \texttt{SimplexLoRA} matches or outperforms standard \texttt{LoRA} and recent adaptive baselines at the same parameter count and inference cost, with the largest gains in the low-rank regime.

\section*{Limitations}\label{sec:limitations}
     The limitations of the proposed strategy are as follows. First, \texttt{SimplexLoRA} introduces three additional hyperparameters: the number of iterations between simplex projection steps \(T\), the total number of such projections \(K\), and the weight decay coefficient used for regularization; the ablation in Section~\ref{sec:ablation_study} indicates that the method is robust within a broad range but does not eliminate the need to set these values.
     Second, the lack of experiments on models substantially larger than 8B parameters limits the generalizability of our findings. Third, while the per-task averages over three seeds consistently favor \texttt{SimplexLoRA}, on a few individual tasks (e.g., CoLA and RTE in the out-of-the-box DeBERTaV3 comparison) standard \texttt{LoRA} remains competitive or slightly better, indicating that the benefit of adaptive rank allocation is task-dependent.
     Fourth, a fundamental ambiguity remains during training: a small gradient norm in a layer may indicate either that the weights are well-optimized or that the gradients have vanished by the time they reach that layer, which can in principle bias the learned importance weights.

    On the ethical side, we see no specific risks beyond those already inherent to fine-tuning large language models: this work studies a parameter-efficient adaptation procedure, introduces no new data sources, uses only publicly available benchmarks with no proprietary or personal data and no human subjects, and reduces rather than increases the compute and memory footprint of fine-tuning.

\section*{Acknowledgments}
    The work of A. Beznosikov was supported by the Russian Science Foundation (project no. 21-71-30005-P).

\end{mainpart}

\begin{appendixpart}

\section{\texttt{SimplexLoRA}: theoretical motivation}\label{appendix:theory}
    This section presents a heuristic argument that motivates the simplex-constrained reparameterization of Section~\ref{sec:simplex_lora}. We do not claim a formal optimization guarantee in the non-convex setting; the goal is to clarify why importance weights on a simplex are a natural device for redistributing adapter capacity.

    Consider a loss function \(\mathcal{L}(\displaystyle \mW^1, \dots, \displaystyle \mW^n)\) for a machine learning model with \(n\) layers. In the context of fine-tuning an LLM using the \texttt{LoRA} algorithm, the optimization problem can be formulated as
    \[
        \min_{\mathcal{A}, \mathcal{B}} \left\{ \mathcal{L}(\mathcal{A}, \mathcal{B}) := \mathcal{L}(\displaystyle \mA^1 \displaystyle \mB^1, \dots, \displaystyle \mA^n \displaystyle \mB^n) \right\},
    \]
    where \(\mathcal{A}\) and \(\mathcal{B}\) denote the sets of all \texttt{LoRA} adapters.
    To generalize this formulation, we introduce a non-negative scaling factor \(\omega_i \geq 0\) for each layer \(i\), which can be interpreted as rescaling the adapter matrix \(\displaystyle \mA^i\) by \(\omega_i\). For the purposes of the present derivation, we constrain the vector \(\omega := (\omega_1, \dots, \omega_n)\) to lie on the probability simplex \(\Delta_{n-1}\), i.e., \(\sum_{i=1}^n \omega_i = 1\); the \(n\)-scaled simplex used in the algorithm (Section~\ref{sec:simplex_lora}) is the same object up to a factor of \(n\), chosen so that the uniform vector \(\omega = \mathbf{1}\) recovers standard \texttt{LoRA}. The resulting optimization problem is then
    \[
        \min_{\mathcal{A}, \mathcal{B}} \left\{ \mathcal{L}(\omega_1 \displaystyle \mA^1 \displaystyle \mB^1, \dots, \omega_n \displaystyle \mA^n \displaystyle \mB^n) \right\}.
    \]

    Note that this reparameterization does not alter the expressive power of the model, as the scaling factors can be absorbed into the adapter matrices.

    A classical simplifying assumption in the optimization literature is that \(\mathcal{L}\) is convex \citep{khaled2023unified} with respect to all parameters \(\{\displaystyle \mW^1, \dots, \displaystyle \mW^n\}\). For deep networks this is generally false, but convex analysis is still commonly used as a guiding lens: empirical studies report local convexity along training trajectories \citep{kleinberg2018alternative, zhou2019sgd}, and many non-convex results build on convex foundations \citep{hinder2020near, fatkhullin2022sharp}. We adopt this assumption only to derive the upper bound below; the resulting prescription on \(\omega\) is then transferred back to the non-convex setting as a design heuristic, not as a guarantee.

    Therefore, if we take \(\mathcal{A}^i \cdot \mathcal{B}^i := \{0, \dots, \displaystyle \mA^i \displaystyle \mB^i, \dots, 0\}\), that is, we fine-tune only the \(i\)-th layer, then, by the convexity of \(\mathcal{L}\) (Jensen's inequality), we obtain: 
    
    \begin{align*}
        &\min_{\mathcal{A}, \mathcal{B}} \left\{ \mathcal{L}(\mathcal{A}, \mathcal{B}) \right\} 
        \\&=
        \min_{\mathcal{A}, \mathcal{B}} \left\{
        \mathcal{L}(\omega_1 \displaystyle \mA^1 \displaystyle \mB^1, \dots, \omega_n \displaystyle \mA^n \displaystyle \mB^n) \right\}
        \\&=
        \min_{\mathcal{A}, \mathcal{B}} \left\{ \mathcal{L}\left( \sum_{i=1}^n \omega_i \cdot \{0, \dots, \displaystyle \mA^i \displaystyle \mB^i, \dots, 0\} \right) \right\}
        \\&\leq
        \min_{\mathcal{A}, \mathcal{B}} \left\{ \sum_{i=1}^n \omega_i \mathcal{L}\left(  \mathcal{A}^i \cdot \mathcal{B}^i \right) \right\}
        \\&=
        \sum_{i=1}^n \omega_i \min_{\displaystyle \mA^i, \displaystyle \mB^i} \left\{ \mathcal{L}(0, \dots, \displaystyle \mA^i \displaystyle \mB^i, \dots, 0) \right\}.
    \end{align*}
    Thus, the original fine-tuning problem is upper-bounded by a weighted sum of $n$ single-layer fine-tuning problems. Minimizing this upper bound over $\omega \in \Delta_{n-1}$ allocates weight to the layer with the smallest single-layer minimum: the $i$-th coordinate of the resulting $\omega$ exceeds the $j$-th if and only if
    \begin{align*}
    \min_{\mathcal{A}^i, \mathcal{B}^i} \mathcal{L}(\mathcal{A}^i \cdot \mathcal{B}^i) \leq \min_{\mathcal{A}^j, \mathcal{B}^j} \mathcal{L}(\mathcal{A}^j \cdot \mathcal{B}^j),
    \end{align*}
    since otherwise swapping their positions would yield a smaller upper bound. Reading this in reverse, components of $\omega$ rank the layers by their single-layer "fitting potential". \texttt{SimplexLoRA} uses this ranking as a proxy for layer importance and reallocates adapter rank accordingly: well-adapted layers receive higher ranks, weaker ones lower. We emphasize that this argument bounds the loss, not the rank distribution itself, so the connection to optimal ranks remains a heuristic; its empirical effectiveness is reported in Section~\ref{sec:simplex_lora} and the ablation in Section~\ref{sec:ablation_study}.

    Note that to prevent \(\omega\) from degenerating into a one-hot vector, which is a common issue in simplex-constrained optimization where a single component dominates, one can apply regularization. In our implementation, we use a standard Euclidean regularization term to encourage \(\omega\) to remain closer to a uniform distribution across all components.


\section{Adapter Reinitialization}
\label{appendix:reinit}
    To ensure that the reinitialized adapters do not introduce immediate perturbations to the model's output, we follow the standard \texttt{LoRA} initialization scheme.
    
    $\bullet$ $\mA^i$ initialized using the Kaiming uniform distribution~\citep{he2015delving} to preserve the variance of activations:
    $
        \mA^i \sim U\left(-d_{in}^{-0.5}, d_{in}^{-0.5}\right),
    $

    $\bullet$ $\mB^i$ initialized to zero.

    This combination ensures that the initial product $\mA^i \mB^i = \mathbf{0}$, maintaining the model's performance at the start of the final training stage while providing a stable gradient flow for the new parameters.

\section{Simplex Projections}
    While softmax and weighted softmax transformations are commonly used for projecting onto the simplex, we observed severe performance degradation during training when applying these methods. Their behavior is highly sensitive to temperature scaling: low temperatures tend to produce nearly one-hot distributions, whereas high temperatures result in almost uniform outputs. Such degeneracies are particularly problematic in our setting, where the goal is to efficiently allocate the rank budget across multiple adapters without inducing excessive sparsity or uniformity. For these reasons, we utilize the standard Euclidean regularization in the optimization problem formulation \eqref{eq:simplex_lora}, rather than alternatives such as KL divergence.

\section{Experimental Details}\label{appendix:exp_comp}
    The core algorithm is implemented in PyTorch using the Huggingface library. All experiments are conducted on NVIDIA A100-SXM4-80GB GPUs. Following standard \texttt{LoRA} practice, we use the scaling factor $\alpha / r$ with $\alpha=32$ and dynamically updated ranks $r_i^{\text{new}}$ \citep{hu2021lora}. To support simplex projection and dynamic rank reallocation, we employ a modified AdamW optimizer \citep{loshchilov2017decoupled} with specialized update steps for $\omega$, rank resizing, and a custom learning rate schedule.
    
    Training and optimization hyperparameters for each experimental setup are summarized in Tables~\ref{tab:hp_out_of_box}, \ref{tab:hp_deberta}, \ref{tab:hp_llama}, \ref{tab:hp_qwen}, and \ref{tab:hp_bart}. The LR warm-up values reported in these tables refer to the learning-rate scheduler and are not \texttt{SimplexLoRA} rank-search hyperparameters. For the first table, the learning rate was either adopted from the original publications or selected via a small grid search. For the DeBERTa, Llama, and Qwen setups, the optimal configuration for each task was determined via an independent grid search over the specified hyperparameter ranges. For GSM8K, we use few-shot chain-of-thought prompting with flexible answer extraction during evaluation. All the results were averaged across 3 random seeds.

\begin{braintablebox}
\captionsetup{type=table}
\centering
\caption{Hyperparameters for out-of-the-box comparison (Table~\ref{tab:glue_all_adapters}).}
\label{tab:hp_out_of_box}
\footnotesize
\setlength{\tabcolsep}{3pt}
\begin{tabular}{p{0.42\linewidth} | p{0.52\linewidth}}
\toprule
\textbf{Hyperparameter} & \textbf{Value} \\
\midrule
Optimizer & AdamW \\
$\beta_1, \beta_2$ & $0.9, 0.999$ \\
Model learning rate & Tuned (PEFT) \\
Learning rate ($\omega$) & $5$ \\
LR warm-up & $30$ \\
Reg. weight decay & 0 \\
Dropout & $0.05$ \\
Grad. accumulation & $6$ \\
Training budget & $512$ iterations \\
Batch size & $32$ \\
Scheduler & Linear \\
\midrule
Target layers & Query, Key, Value \\
\bottomrule
\end{tabular}
\end{braintablebox}

\begin{braintablebox}
\captionsetup{type=table}
\centering
\caption{Hyperparameters for DeBERTaV3\textsubscript{base} experiments (Table~\ref{tab:llama_deberta_glue}).}
\label{tab:hp_deberta}
\footnotesize
\setlength{\tabcolsep}{3pt}
\begin{tabular}{p{0.42\linewidth} | p{0.52\linewidth}}
\toprule
\textbf{Hyperparameter} & \textbf{Value / Grid} \\
\midrule
Optimizer & AdamW \\
$\beta_1, \beta_2$ & $0.9, 0.999$ \\
Model learning rate & $\{3, 5, 8, 10, 30, 50, 80, 100\} \!\times\! 10^{-5}$ \\
Learning rate ($\omega$) & $5$ \\
LR warm-up & $\{0, 20, 30, 50, 70, 100, 200\}$ \\
Reg. weight decay & $\{0, 10^{0}, 10^{-1}, \dots, 10^{-6}\}$ \\
Dropout & $0.05$ \\
Grad. accumulation & $6$ \\
Training budget & task-specific \\
Batch size & task-specific \\
Scheduler & Two-phase (Linear) \\
\midrule
Target layers & All linear layers \\
\midrule
\multicolumn{2}{l}{\textbf{Task-specific (\# epochs / batch size)}} \\
MNLI / SST-2 / QNLI / QQP & (7 / 32), (10 / 32), (5 / 32), (5 / 32) \\
CoLA / RTE / MRPC & (15 / 64), (50 / 64), (30 / 64) \\
\bottomrule
\end{tabular}
\end{braintablebox}

\begin{braintablebox}
\captionsetup{type=table}
\centering
\caption{Hyperparameters for Llama~3.1~8B experiments (Table~\ref{tab:llama_deberta_glue}).}
\label{tab:hp_llama}
\footnotesize
\setlength{\tabcolsep}{3pt}
\begin{tabular}{p{0.42\linewidth} | p{0.52\linewidth}}
\toprule
\textbf{Hyperparameter} & \textbf{Value / Grid} \\
\midrule
Optimizer & AdamW \\
$\beta_1, \beta_2$ & $0.9, 0.999$ \\
Model learning rate & $\{3, 5, 8, 10, 30, 50, 80, 100\} \!\times\! 10^{-5}$ \\
Learning rate ($\omega$) & 5 \\
LR warm-up & 50 \\
Reg. weight decay & $\{0, 10^{0}, 10^{-1}, \dots, 10^{-6}\}$ \\
Dropout & $0.05$ \\
Grad. accumulation & $1$ \\
Training budget & $1024$ iterations \\
Batch size & $16$ \\
Scheduler & Two-phase (Cosine) \\
\midrule
Target layers & All linear layers \\
\bottomrule
\end{tabular}
\end{braintablebox}

\begin{braintablebox}
\captionsetup{type=table}
\centering
\caption{Hyperparameters for Qwen3-8B experiments (Table~\ref{tab:qwen_bart}).}
\label{tab:hp_qwen}
\footnotesize
\setlength{\tabcolsep}{3pt}
\begin{tabular}{p{0.42\linewidth} | p{0.52\linewidth}}
\toprule
\textbf{Hyperparameter} & \textbf{Value / Grid} \\
\midrule
Optimizer & AdamW \\
$\beta_1, \beta_2$ & $0.9, 0.999$ \\
Model learning rate & $\{3, 5, 8, 10, 30, 50, 80, 100\} \!\times\! 10^{-5}$ \\
Learning rate ($\omega$) & 5 \\
LR warm-up & 350 \\
Reg. weight decay & $\{0, 10^{0}, 10^{-1}, \dots, 10^{-6}\}$ \\
Dropout & $0.05$ \\
Grad. accumulation & $1$ \\
Training budget & $3500$ iterations \\
Batch size & $4$ \\
Scheduler & Two-phase (Linear) \\
\midrule
Target layers & All linear layers \\
\bottomrule
\end{tabular}
\end{braintablebox}

\begin{braintablebox}
\captionsetup{type=table}
\centering
\caption{Hyperparameters for BART-large experiments (Table~\ref{tab:qwen_bart}).}
\label{tab:hp_bart}
\footnotesize
\setlength{\tabcolsep}{3pt}
\begin{tabular}{p{0.42\linewidth} | p{0.52\linewidth}}
\toprule
\textbf{Hyperparameter} & \textbf{Value / Grid} \\
\midrule
Optimizer & AdamW \\
$\beta_1, \beta_2$ & $0.9, 0.999$ \\
Model learning rate & Tuned (PEFT) \\
Learning rate ($\omega$) & $5$ \\
LR warm-up & $50$ \\
Reg. weight decay & $10^{-4}$ \\
Model weight decay & $10^{-1}$ \\
Dropout & $0.05$ \\
Grad. accumulation & $16$ \\
Training budget & $15$ epochs \\
Batch size & $8$ \\
Scheduler & Two-phase (Linear) \\
\midrule
Beam size & $8$ \\
Max source / target length & $256 / 160$ \\
Target layers & All linear layers \\
\bottomrule
\end{tabular}
\end{braintablebox}

\section{Use of AI Assistants}\label{appendix:ai_assistants}
    During the preparation of this work, the authors used a LLM to assist with several tasks. Specifically, the LLM was employed for code generation to create data visualization scripts, debugging existing code, and for proofreading the paper to improve grammatical accuracy and stylistic consistency. After using this tool, the authors reviewed and edited the content as needed for the final publication.

\end{appendixpart}
\end{document}

%% file: math_commands.tex
\usepackage{amsmath,amsfonts,bm}

\def\eqref#1{equation~\ref{#1}}

\def\1{\bm{1}}

\def\va{{\bm{a}}}
\def\vb{{\bm{b}}}

\def\mA{{\bm{A}}}
\def\mB{{\bm{B}}}

\def\mI{{\bm{I}}}

\def\mP{{\bm{P}}}
\def\mQ{{\bm{Q}}}

\def\mW{{\bm{W}}}

\DeclareMathAlphabet{\mathsfit}{\encodingdefault}{\sfdefault}{m}{sl}
\SetMathAlphabet{\mathsfit}{bold}{\encodingdefault}{\sfdefault}{bx}{n}

\newcommand{\R}{\mathbb{R}}

